\documentclass[letterpaper, 10 pt, conference]{ieeeconf}
\IEEEoverridecommandlockouts
\usepackage{epsfig}
\usepackage{multirow}
\usepackage{subcaption}
\usepackage[nolist]{acronym}
\begin{acronym}
\acro{vo}[VO]{Visual Odometry}
\acro{cnn}[CNN]{Convolutional Neural Network}
\acro{rnn}[RNN]{Recurrent Neural Network}
\acro{mdn}[MDN]{Mixture Density Network}
\acro{lstm}[LSTM]{Long Short-Term Memory}
\acro{mlp}[MLP]{Multilayer Perceptron}
\acro{imu}[IMU]{Inertial Measurement Unit}
\acro{rpe}[RPE]{Relative Pose Error}
\acro{ekf}[EKF]{Extended Kalman Filter}
\acro{mse}[MSE]{Mean Squared Error}
\end{acronym}

\title{\LARGE \bf
Multi-Camera Sensor Fusion for Visual Odometry using Deep Uncertainty Estimation
}

\author{Nimet Kaygusuz, Oscar Mendez, Richard Bowden

\thanks{All authors are with the University of Surrey.
        {\tt\small \{n.kaygusuz, o.mendez, r.bowden\}@surrey.ac.uk}}
}

\begin{document}

\maketitle
\thispagestyle{empty}
\pagestyle{empty}

\begin{abstract}

\ac{vo} estimation is an important source of information for vehicle state estimation and autonomous driving. Recently, deep learning based approaches have begun to appear in the literature. However, in the context of driving, single sensor based approaches are often prone to failure because of degraded image quality due to environmental factors, camera placement, etc. To address this issue, we propose a deep sensor fusion framework which estimates vehicle motion using both pose and uncertainty estimations from multiple on-board cameras.

We extract spatio-temporal feature representations from a set of consecutive images using a hybrid CNN - RNN model. We then utilise a \acf{mdn} to estimate the 6-DoF pose as a mixture of distributions and a fusion module to estimate the final pose using \ac{mdn} outputs from multi-cameras.

We evaluate our approach on the publicly available, large scale autonomous vehicle dataset, nuScenes. The results show that the proposed fusion approach surpasses the state-of-the-art, and provides robust estimates and accurate trajectories compared to individual camera-based estimations.

\end{abstract}

\section{INTRODUCTION}

Perception of the environment via sensor systems is a major challenge in the quest to realise autonomous driving. To ensure safety, accurate and robust systems are required that can operate within any environment. One of the most fundamental components of any autonomous system is ego-motion estimation which can be used for both vehicle control and automated driving.

In recent years, considerable research effort has been invested into ego-motion estimation using on-board sensors \cite{chen2020survey}. It is common to employ various types of complementary sensors, e.g., IMU, GPS, LiDAR and cameras, to provide accurate motion for vehicle state estimation systems.

Cameras are a popular choice as they are cost effective, already installed on most modern vehicles and we know that vision is key to human driving ability. However, the accuracy of vision-based motion estimation is affected by various internal and external factors. Even if we assume that camera calibration is known and correct, the quality of the captured images are dependent on environmental factors such as occlusions, lighting and weather conditions. Considering that \acf{vo} works by examining the change in image pixel motion between consecutive frames, robust and distinctive features are crucial for its performance. This means that the quality of estimated motion is dependent upon the quality of images, which in turn is dependent upon the type and structure of the scene.

\begin{figure}[t]
\begin{center}
   \includegraphics[width=1.0\linewidth]{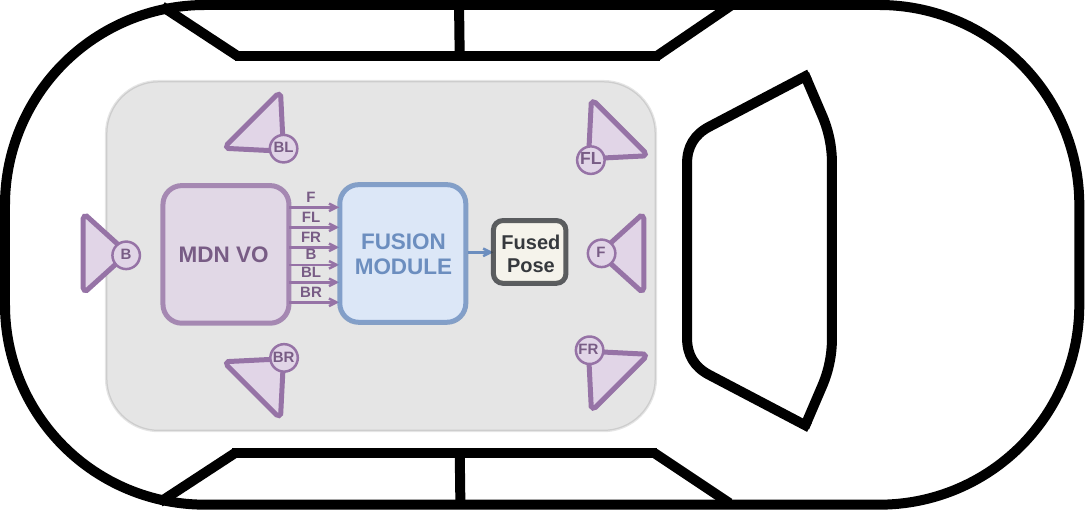}
\end{center}
   \caption{An overview of the proposed multi-camera fusion approach.}
\label{fig:vehicle}
\end{figure}

Robustness can be provided in many ways, but for vehicle state estimation, sensor fusion techniques are often employed. Many successful sensor fusion frameworks employ a Kalman filter (or variant) \cite{caron2006gps, wang2018gnss}. Kalman filters are designed to handle the sensor measurement noise and allow the combination of complementary sensors. However, measurement statistics and system dynamics needs to be known for the filter to operate successfully. As image quality is dependent upon environmental factors, fusion of multiple cameras in a VO system requires estimates of sensor reliability or measurement noise and this implies implicit knowledge about what environmental factors lead to degradation. 

Motivated by recent developments in deep learning and its applications to autonomous vehicles, we propose a learning based sensor fusion framework which fuses multiple \ac{vo} estimates from multiple cameras, taking into account the predicted confidence of each sensor. 

We utilise a hybrid \ac{cnn} - \ac{rnn} architecture that extracts spatio-temporal features from images. We then employ a \acf{mdn} which predicts the probability distributions of motion for each camera independently. This approach allows our model to learn the confidence/uncertainty of motion for each camera. Independent motion estimates from multiple arms of the network are then combined in a neural fusion module. An overview of our approach can be seen in Figure~\ref{fig:vehicle}.

We evaluate the proposed fusion approach on nuScenes \cite{caesar2020nuscenes} which is a large scale, public, autonomous driving dataset. We report quantitative and qualitative \ac{vo} results. Our experiments show that, overall, the proposed fusion approach can provide more accurate estimations than individual cameras alone. Furthermore, our model surpasses the two state-of-the-art \ac{vo} algorithms and baseline \ac{ekf} based fusion.

The contributions of this work can be summarised as:
\begin{itemize}
    \item We utilise a deep neural network approach to estimate both vehicle motion and the confidence in that estimation, from multiple monocular cameras. 
    \item We propose a deep fusion framework that can intelligently fuse multiple motion estimates incorporating the pose and uncertainties in prediction.
    \item We demonstrate state-of-the-art performance on a large-scale autonomous driving dataset and evaluate performance under different weather and lighting conditions. 
\end{itemize}

The rest of this paper is structured as follows: In Section~\ref{sec:related-work} we discuss relevant related work. We then introduce the proposed learning based fusion model in Section~\ref{sec:methodology}. We share the experimental setup and report our results in Section~\ref{sec:experiments}. Finally, we conclude the paper in Section~\ref{sec:conclusion} by discussing our findings.

\section{RELATED WORK}
\label{sec:related-work}
This paper focuses on estimating vehicle motion from multiple on-board cameras. \ac{vo} can be defined as estimating the motion of a camera by examining the motion of features between consecutive image frames. Traditionally, there are two families of approaches to estimating vehicle motion via vision sensors, namely direct methods and feature based methods. Direct methods \cite{newcombe2011dtam, engel2017direct} track the changes in pixel intensities to estimate the camera motion. Feature based methods \cite{davison2007monoslam, mur2015orb} use handcrafted features to detect salient patterns and track their displacement on the image plane in order to estimate the camera motion \cite{scaramuzza2011visual}. Even though both methods show promising results, they both have their short comings. Feature based methods are prone to failure in low texture environments or with insufficient lighting, while direct methods tend to fail at high velocities and are sensitive to photometric changes. 
    
More recently, deep learning based models have been applied to the \ac{vo} task \cite{wang2017deepvo, li2018undeepvo, zhan2018unsupervised}. One of the main advantages of these models is not requiring handcrafted features, due to their ability to learn meaningful feature representations for the task they were trained upon. Compared to classical \ac{vo} approaches, they are more robust in low illumination and texture-less environments \cite{chen2020survey}.
In \cite{mohanty2016deepvo}, Mohanty et al. proposed estimating motion via monocular camera from consecutive frames using \ac{cnn}s which were trained in a supervised manner. Wang et al. \cite{wang2017deepvo, wang2018end} expanded this approach and introduced the use of \ac{rnn} to model the temporal dependencies between frames. In this work, we employ a similar supervised approach to estimate motion for individual cameras. However, instead of directly regressing a 6-DoF pose, we estimate the motion as a mixture of Gaussians using \ac{mdn}'s. This enables our model to estimate pose and its uncertainty together which form the input of our fusion framework.

One of the main shortcomings of supervised deep learning approaches is their need to be trained on large scale datasets. To address this issue, unsupervised \ac{vo} methods have been studied. Zhou et al. \cite{zhou2017unsupervised} proposed learning depth maps and camera pose change together with view synthesis. Li et al. \cite{li2018undeepvo} utilised stereo images which enabled their network to recover the global scale. More recently, Yang et al. \cite{yang2020d3vo} proposed learning depth, pose and photometric uncertainty in an unsupervised manner, where they used photometric uncertainties to optimise the \ac{vo} estimates. Although unsupervised \ac{vo} methods have the potential of exploiting unlabelled data, these approaches do not have the necessary robustness for vehicle state estimation.

To improve the robustness of vehicle motion estimation, sensor fusion techniques have been employed \cite{parra2011visual, lynen2013robust, chen2019selective}. However, most of the sensor fusion studies focus on using \ac{vo} as a complementary source of information. Parra et al. \cite{parra2011visual} use \ac{vo} when GPS fails. Lynen et al. fuse \ac{imu} and vision sensors using a Kalman filter. Chen et al. \cite{chen2019selective} proposed a fusion framework which selectively fuses a monocular camera and \ac{imu} by eliminating the corrupted sensor measurements. 

To perform more reliably than any individual sensor, sensor fusion techniques require measurement uncertainty. But building an accurate uncertainty model for sensors is a difficult task \cite{fayyad2020deep}. In addition to the sensor imperfections, adverse environmental conditions have a significant effect on sensor output. For example, visual sensors, such as colour cameras, are particularly sensitive to environmental factors such as direct sunlight, low texture areas or poor lighting. Thus, semantic understanding of the scene is required in order to estimate the accuracy of vision sensors. In this work we model the measurement uncertainty using an \ac{mdn}. By estimating a probability distribution over the motion for each camera, we can robustly fuse multiple sensors together in a single multi-stream neural network to overcome the possible failure of individual sensors.  

\begin{figure*}[t]
\vspace{0.09cm}
\begin{center}
   \includegraphics[width=0.70\linewidth]{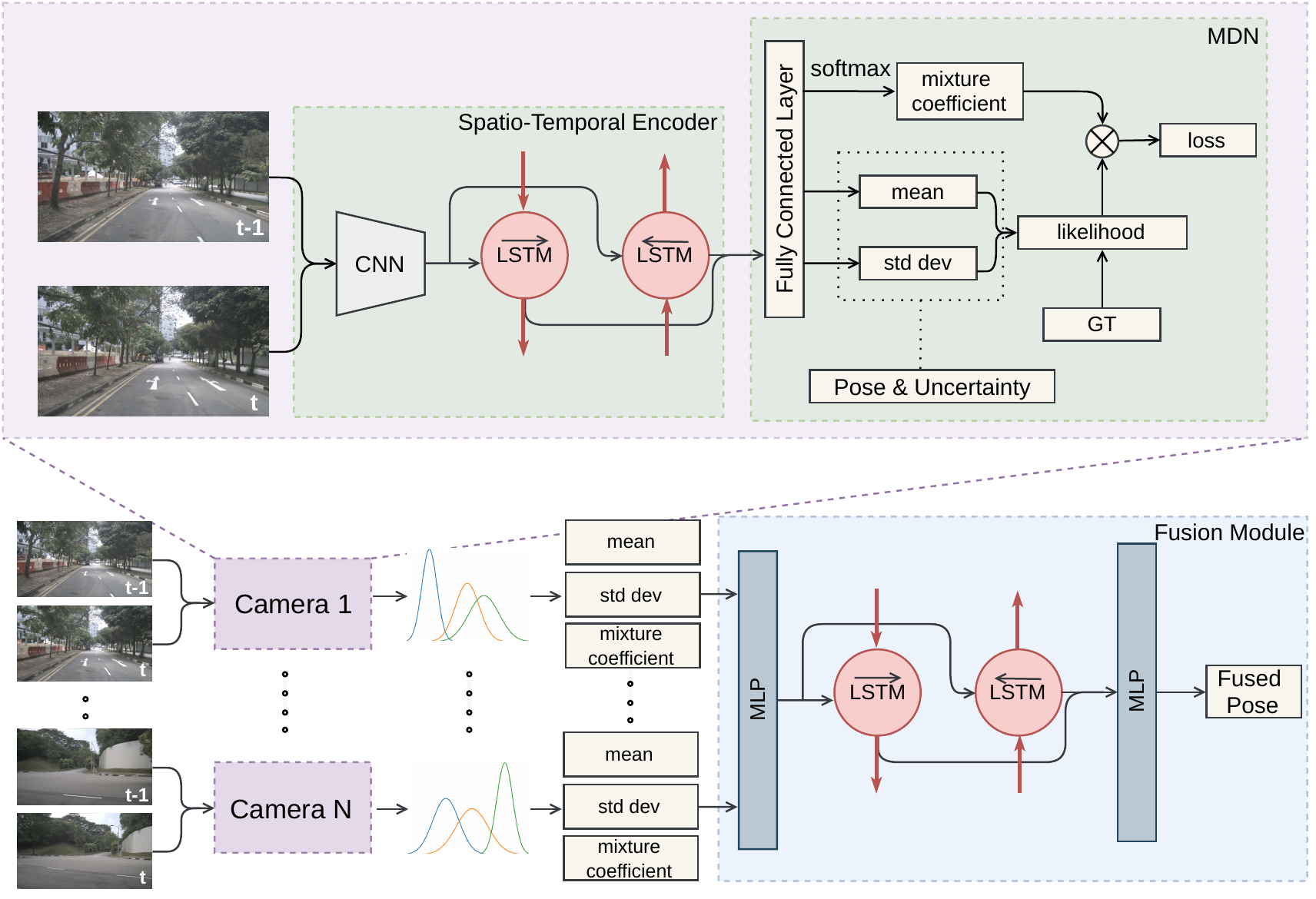}
\end{center}
   \caption{An overview of the proposed multi-camera sensor fusion for \ac{vo}. The fusion module uses pose and uncertainty estimations from the \ac{mdn} module.}
\label{fig:architecture}
\end{figure*}

\section{METHODOLOGY}
\label{sec:methodology}
In this section, we introduce the proposed multi-camera fusion approach for estimating \ac{vo}. Our approach starts by acquiring video streams from multiple cameras that are mounted to the vehicle with different positions \& orientations. Given video streams $V^{1:N} = \{I_0^n, ..., I_T^n\}^{1:N}$ from $N$ cameras with $(T+1)$ number of frames, our model predicts the 6-DoF relative poses $Y = \{y_{(0, 1)}, ..., y_{(T-1,T)}\}$ between consecutive time steps.

For each camera, we first extract spatio-temporal feature representations using a hybrid \ac{cnn} - \ac{rnn} architecture (Section~\ref{sec:ste}). We then pass these representations to an \ac{mdn} module (Section~\ref{sec:mdn}), which predicts a mixture of Gaussian distributions for the vehicle pose. By running multiple streams of \ac{vo} through different arms of the network simultaneously, we then combine the pose distributions coming from multiple cameras in a deep fusion module (Section~\ref{sec:sf}). This produces a final pose estimate. An overview of the proposed multi-camera fusion approach is visualised in Figure~\ref{fig:architecture}.

\subsection{Spatio-Temporal Encoder}
\label{sec:ste}
Estimating motion from video requires both spatial and temporal understanding. Temporal modelling can be further divided into short term and long term tracking. While we are interested in short term changes between consecutive frames, i.e. relative pose, we also want to model the long term motion of the vehicle to help produce consistent estimates. 
In this work, we model spatial and short-term temporal representations using a CNN backbone similar to those used for optical flow estimation~\cite{dosovitskiy2015flownet}, which is a related task. Given a consecutive image pair $[I_{t-1}^n, I_t^n]$ from the $n^{th}$ camera, we extract the motion features, $f_t^n$ as:

\begin{equation}
    f_t^n = \mathrm{CNN}([I_{t-1}^n, I_t^n])
\end{equation}

where $[.]$ is the concatenation operation over the image colour channels. The \ac{cnn} network is composed of $9$ convolutional layers. After each convolutional layer, we include batch normalisation \cite{ioffe2015batch}, ReLU \cite{maas2013rectifier} and Dropout \cite{srivastava2014dropout}.

To model the longer term vehicle motion, we feed the features from the  \ac{cnn}, $f_{1:T}^n$, to an \ac{rnn} module. At each time step $t$, the \ac{rnn} produces temporally enhanced representations $r_t^n$ as:

\begin{equation}
    r_t^n = \mathrm{RNN}(f_t^n,h_{t-1}^n)
\end{equation}

where $f_t^n$ and $h_{t-1}^n$ are visual features and the previous hidden states at time $t$ for the $n^{th}$ camera, respectively. We initialise the hidden state as all zeros for $t = 1$. In this work we employed a bi-directional \ac{lstm} \cite{hochreiter1997long} and used a small sliding window of past $5$ frames. However, any \ac{rnn} architecture can be utilised with our approach.

\subsection{Mixture Density Network}
\label{sec:mdn}
Learning based approaches commonly suffer from the problem of regression to the mean, which can be considered as approximating the conditional average of the output. To address this issue, we utilise an \ac{mdn}, which estimates a mixture of distributions of the 6-DoF relative poses.

We choose to construct our mixture model from Gaussian distributions to model the vehicle's pose. For each camera $n$, our model estimates a mixture model, $\mathcal{G}^n$, with $M$ components ($5$ in our experiments) at every time step $t$, which can be notated as: 

\begin{equation}
\mathcal{G}_t^n = \{(\alpha^1_t\mathcal{N}_t(\mu^1, \sigma^1))^n, ..., (\alpha^M_t \mathcal{N}_t(\mu^M, \sigma^M))^n\}
\label{eq:mixparams}
\end{equation}

where $\mu$, $\sigma$ and $\alpha$ represent the mean, standard deviation and mixture coefficients, respectively. This approach enables us to model the variances in the output and provide an estimate of how confident our model is in its estimation.

Given the \ac{lstm} outputs of the $n^{th}$ camera, $r_t^n$, we estimate the conditional density of the 6-DoF pose, $y_{(t-1,t)}$, for the $i^{th}$ mixture component, $\phi_i (y_{(t-1,t)}^n|r_t^n)$, as: 

\begin{equation}
    \phi \left(y_{(t-1,t)}^n|r_t^n\right) = \frac{1}
    {\left(\sigma_i\left(r_t\right)\right)^n \sqrt{2{\pi}}} 
    e^{\left\{-\frac{{\|y_{(t-1,t)}^n - \left(\mu_i\left(r_t\right)\right)^n\|}^2}
    {2\left(\sigma_i\left(r_t\right)^2\right)^n}\right\}}
\end{equation}

where $(\mu_i\left(r_t\right))^n$ and $(\sigma_i\left(r_t\right))^n$ denote mean and standard deviation of $i^{th}$ distribution, conditioned on the features, $r_t^n$.

The probability density of the 6-DoF pose, is represented as a linear combination of mixture components as:

\begin{equation}
    p\left(y_{(t-1,t)}^n|r_t^n\right) = \sum_{i=1}^{M}
    \left(\alpha_{i}\left(r_t\right)\right)^n \phi_i \left(y_{(t-1,t)}^n|r_t^n\right)
\end{equation}

where $(\alpha_i\left(r_t\right))^n$ is the mixture coefficient which represents the probability of the pose being generated from $i^{th}$ component.

Finding the appropriate weights for our model can be achieved by maximising the likelihood $\mathcal{L}^n$. Thus, we train our model by minimising the negative log likelihood of the ground truth being generated by the mixture distribution as:

\begin{equation}
    E^n = \mathrm{-log}(\mathcal{L}^n)
\end{equation}

We train \ac{mdn} modules independently for each view. Estimating parameters of a mixture of distributions \cite{bishop1994mixture} allows our model to predict the variance of the pose which explicitly represents the \ac{vo} uncertainty.

\subsection{Uncertainty Based Deep Sensor Fusion}
\label{sec:sf}
The fusion module within the network combines the predictions from individual cameras and estimates a final 6-DoF pose transformation. In a similar fashion to using covariance matrices in a Kalman filter to model measurement uncertainties, we use the output of each camera  \ac{mdn} (including the means, standard deviations and mixture coefficients) as input to the fusion module.

We concatenate the mixture model estimates $\{\mathcal{G}_t^1,...,\mathcal{G}_t^N\}$ from $N$ cameras, and project them to a latent space using a \ac{mlp} with dropout and ReLU activation as: 

\begin{equation}
    p_t = \mathrm{MLP}([\mathcal{G}_t^1,...,\mathcal{G}_t^N])
\end{equation}

where $\mathcal{G}_t^n$ is the mixture model parameters of the $n^{th}$ camera at time $t$ (See Equation~\ref{eq:mixparams}). To encourage a temporally consistent final pose estimate, we employ an \ac{rnn} model. We pass \ac{mlp} outputs $p_{0:T}$ to the \ac{rnn} layer and extract temporal representation, $q_t$ as:

\begin{equation}
    q_t = \mathrm{RNN}(p_t, h_{t-1})
\end{equation}

where $h_{t-1}$ is the hidden state from the previous time step. Finally, we employ a final \ac{mlp} layer to estimate the fused 6-DoF relative pose, $z_t$ as:

\begin{equation}
    z_t = \mathrm{MLP}(q_t)
\end{equation}

We train our fusion module using \ac{mse} loss function. To balance the translation and rotation errors we apply a weighted sum and calculate the error as:

\begin{equation}
    \mathcal{E} = \frac{1}{T} \sum^T_{t=1}{||y_{(t-1, t)}^\rho - z_t^\rho||^2_2} + \lambda_{\varphi} {||y_{(t-1, t)}^\varphi - z_t^\varphi||^2_2}
\end{equation}

where $y_{(t-1, t)}^\rho$ and $y_{(t-1, t)}^\varphi$ denote the translation and rotation (Euler angles) of the ground truth relative poses, respectively. $\lambda_{\varphi}$ is an empirically chosen hyper-parameter to weight the rotation errors, which is set to be $100$ in our experiments.

\section{EXPERIMENTS}
\label{sec:experiments}

\subsection{Dataset and Implementation Details}
We evaluate our approach on the recently released nuScenes dataset \cite{caesar2020nuscenes}, which is a large-scale autonomous driving dataset consisting of $1.4$ million camera images containing different types of manoeuvres (velocities ranging from $0$ km/h to $63$ km/h), weather and lighting conditions. It has six cameras attached at different locations \& orientations around the vehicle. The cameras have $70^\circ$ FOV for front and side and $110^\circ$ FOV for rear providing a $360$ degree view. This makes nuScenes an ideal dataset for multi-camera fusion. nuScenes has $850$ driving sequences with available ground truth poses. Among them we eliminate the static scenarios, where the vehicle does not move in the sequence. It should be noted that the remaining scenes already contain static parts of a trajectory which can be easily learned by the model. We split the sequences for training and test, yielding 676 and 100 sequences in each set, respectively.

We implemented our network using the PyTorch deep learning framework \cite{paszke2017automatic}. Adam optimiser \cite{kingma2014adam} is used to train our network with a learning rate of $10^{-3}$ $(\beta_1=0.9, \beta_2=0.999)$. We utilise plateau learning rate scheduling with a patience of $8$ and a decay factor of $0.7$. We use pre-trained FlowNet \cite{dosovitskiy2015flownet} weights to initialise our \ac{cnn} backbone. We use Xavier initialisation \cite{glorot2010understanding} for the remaining parameters.
We train our networks on a machine equipped with an NVIDIA Titan X GPU. The training takes approximately $70$ epochs. At inference, our model runs at $20$ ms per frame on the same machine. Considering nuScenes cameras run at $12$ Hz, the proposed approach meets real time requirements.
To evaluate the performance of our approach, we use the \textit{evo} python package \cite{grupp2017evo} and report \acf{rpe} since it measures the local accuracy of a trajectory, which is a standard way to evaluate \ac{vo} systems.

\subsection{Comparing Fusion Results with Individual Cameras}

In our first set of experiments, we compare the performance of the proposed multi-camera fusion approach against the \ac{vo} estimations from individual cameras. As mentioned in Section~\ref{sec:mdn}, the \ac{mdn} module estimates a mixture of distributions over the 6-DoF pose for each camera. Here, we use the mean of the estimated Gaussians as their \ac{vo} predictions and compare them to our fusion module's estimations (See Section~\ref{sec:sf}) to see the efficacy of the proposed approach compared to using only single camera estimations. 

\begin{table}[hb!]
\centering
\caption{Relative pose errors of monocular \ac{vo} and the proposed fusion approach.}
\begin{tabular}{l|c|c|c}
Camera Views & \multicolumn{1}{c|}{RMSE} & \multicolumn{1}{c|}{Max} & Mean $\pm$ std \\ \hline
FRONT       & 0.077 & 0.531 & 0.049 $\pm$ 0.058 \\
FRONT LEFT  & 0.086 & 0.536 & 0.057 $\pm$ 0.062 \\
FRONT RIGHT & 0.087 & 0.563 & 0.056 $\pm$ 0.066 \\
BACK        & 0.090 & 0.621 & 0.054 $\pm$ 0.070 \\
BACK LEFT   & 0.111 & 0.599 & 0.073 $\pm$ 0.082 \\
BACK RIGHT  & 0.085 & 0.531 & 0.056 $\pm$ 0.064 \\ \hline
\textbf{FUSION} & \textbf{0.045} & \textbf{0.171} & \textbf{0.035 $\pm$ 0.028}
\end{tabular}
\label{table:individual}
\end{table}

\begin{figure*}[!ht]
\begin{subfigure}[b]{0.3\textwidth}
\centering
  \includegraphics[width=1.0\linewidth]{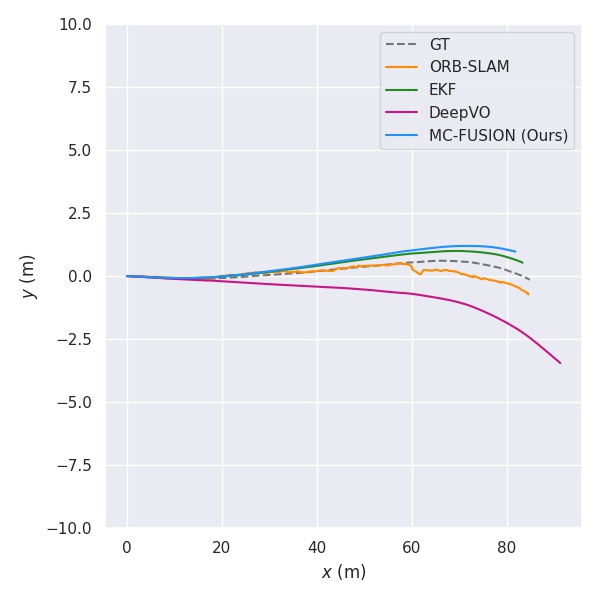}
  \hspace*{0.7cm}
  \includegraphics[width=0.75\linewidth]{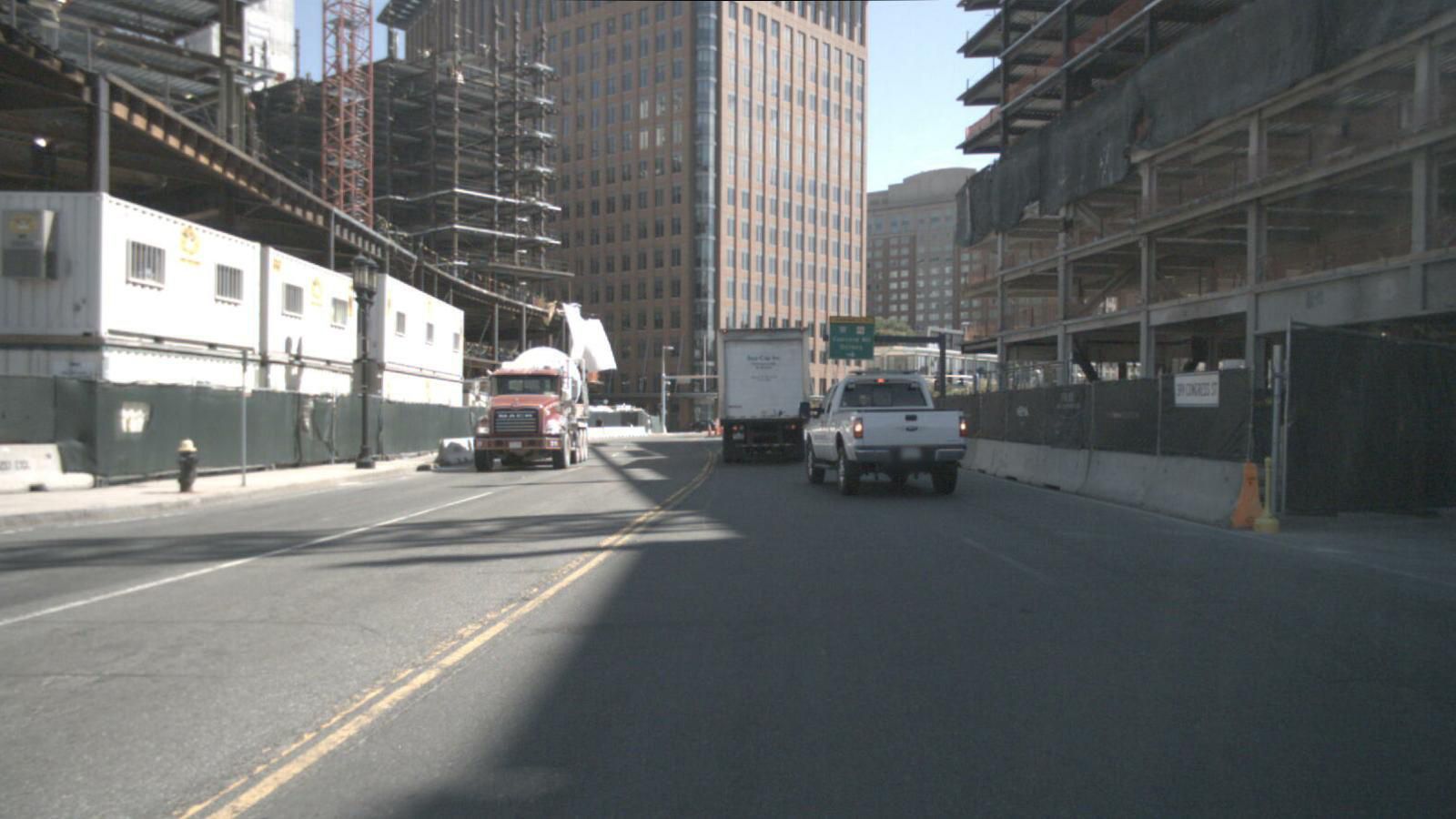}
  \caption{Daylight / scene-0303}
  \label{fig:nuscenesresult:a}
\end{subfigure}
\begin{subfigure}[b]{0.3\linewidth}
\centering
  \includegraphics[width=1.0\linewidth]{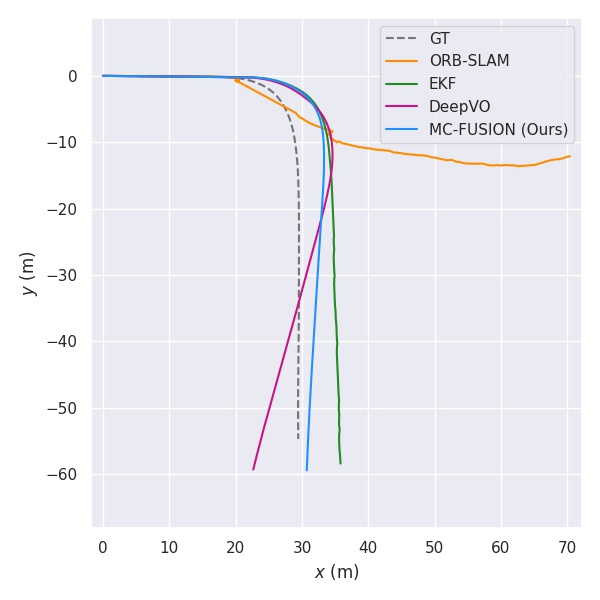}
  \hspace*{0.7cm}
  \includegraphics[width=0.75\linewidth]{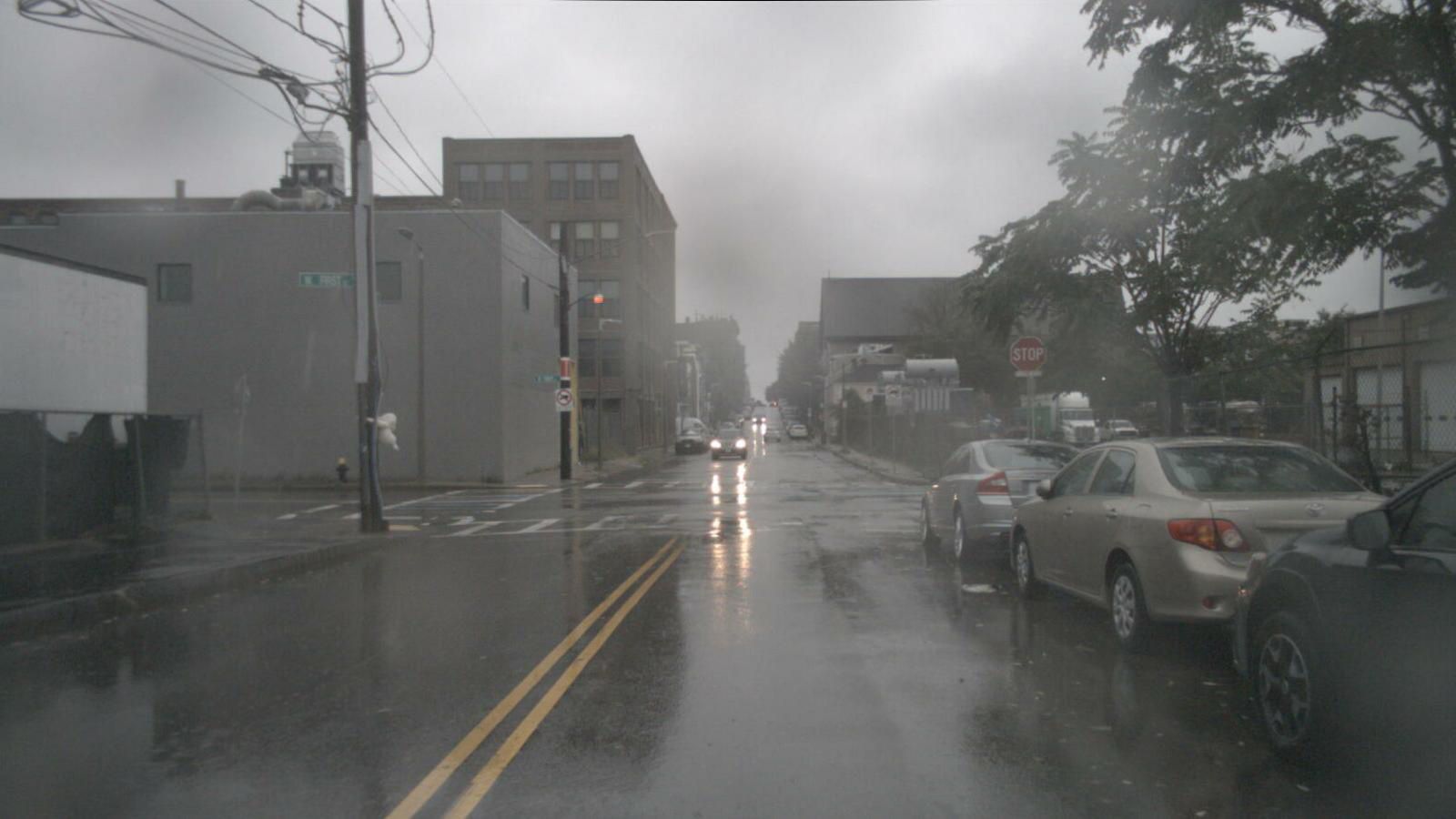}
  \caption{Rain / scene-0570}
  \label{fig:nuscenesresult:b}
\end{subfigure}
\begin{subfigure}[b]{0.3\linewidth}
\centering
  \includegraphics[width=1.0\linewidth]{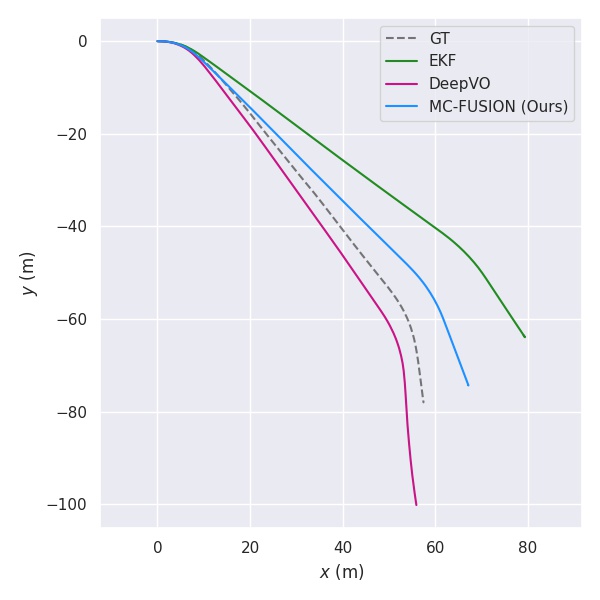}
    \hspace*{0.7cm}
    \includegraphics[width=0.75\linewidth]{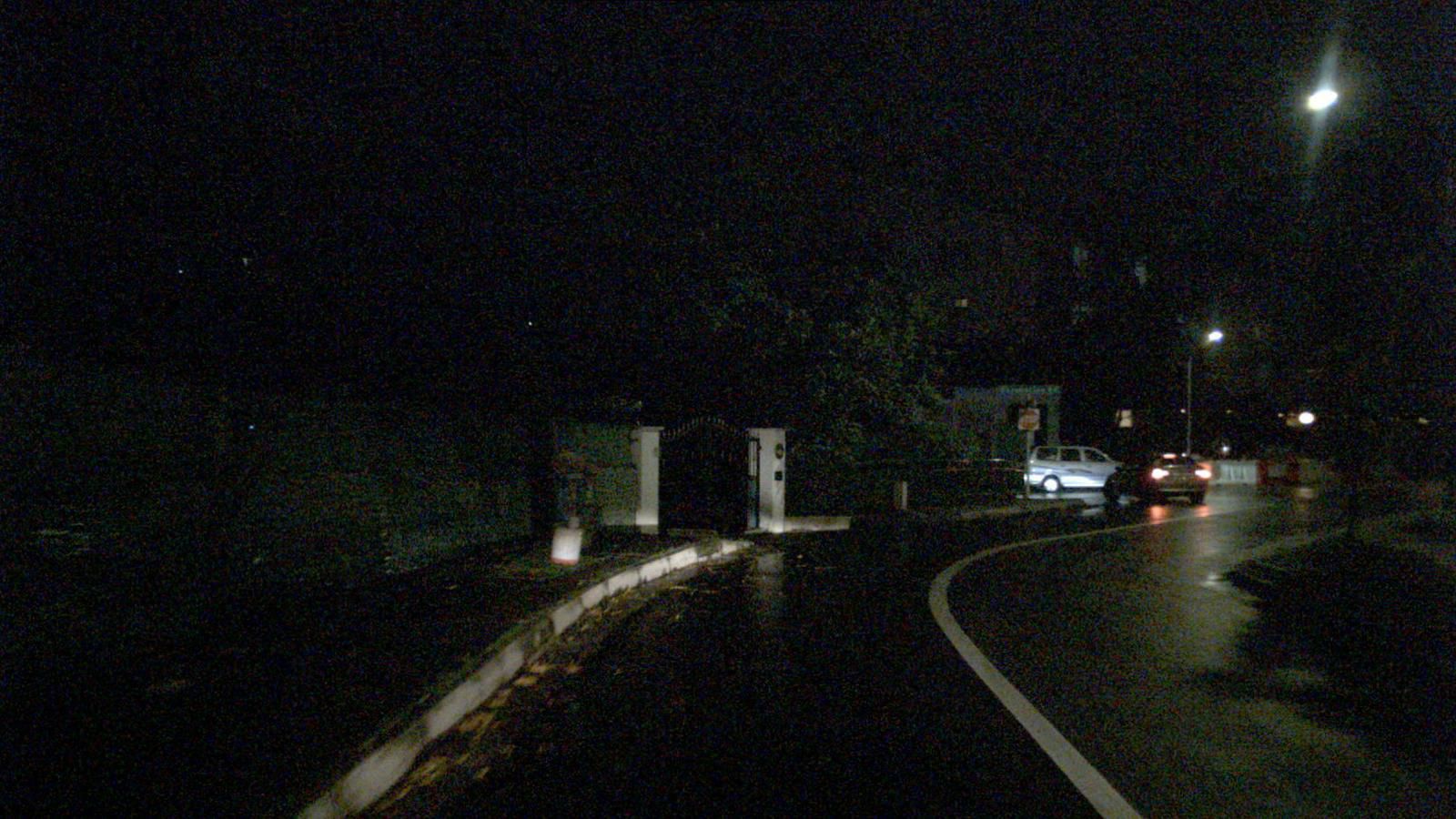}
    \caption{Night / scene-1062}
    \label{fig:nuscenesresult:c}
\end{subfigure}
  \caption{Estimated trajectories (top) and sample frames (bottom) from the corresponding nuScenes sequences.}
\label{fig:nuscenesresult}
\end{figure*}

\begin{table*}[!ht]
\centering
\caption{Relative pose errors on the nuScenes test sequences, categorised by weather/lighting conditions. NR represents sequences where an approach fail to produce results.}
\begin{tabular}{c|ccc|ccc|ccc}

\multirow{2}{*}{} & \multicolumn{3}{c|}{Daylight (64 seq.)} & \multicolumn{3}{c|}{Rain (12 seq.)} & \multicolumn{3}{c}{Night (24 seq.)} \\ \cline{2-10} 
          & RMSE & Max   & Mean $\pm$ std & RMSE & Max   & Mean $\pm$ std  & RMSE & Max  & Mean $\pm$ std \\ \hline
{ORB-SLAM \cite{mur2015orb}} & 0.40 & 3.28  & 0.12 $\pm$ 0.38 & 0.76 & 10.19 & 0.14 $\pm$ 0.74 & NR    & NR     & NR \\
{DeepVO \cite{wang2017deepvo}} & 0.09 & 0.54 & 0.06 $\pm$ 0.07 & 0.07 & 0.39 & 0.05 $\pm$ 0.05 & 0.12 & 0.74 & 0.09 $\pm$ 0.09 \\ \hline
{EKF \cite{moore2016generalized}} & 0.07 & 0.35 & 0.05 $\pm$ 0.05 & 0.06 & 0.36 & 0.04 $\pm$ 0.04 & 0.16 & 0.86 & 0.8 $\pm$ 0.12 \\
\multicolumn{1}{c|}{MC-Fusion (Ours)} & \textbf{0.04} & \textbf{0.15} & \textbf{0.03 $\pm$ 0.02} & \textbf{0.04} & \textbf{0.14} & \textbf{0.03 $\pm$ 0.03} & \textbf{0.07} & \textbf{0.26} & \textbf{0.05 $\pm$ 0.04} \\
\end{tabular}
\label{table:nuscenes}
\end{table*}

Table \ref{table:individual} shows the \ac{vo} performance from six different cameras compared to the result of our fusion model. We use the average \ac{rpe} on the nuScenes test split which consist of $100$ unseen sequences. As can be seen, the fusion results outperform all individual camera based estimations. This is because the proposed fusion model can employ complementary information from the different views. This demonstrates the efficacy of our multi-camera fusion approach and verifies that \ac{vo} accuracy can be improved using the proposed uncertainty-based fusion approach.

\subsection{Comparison against the state-of-the-art}

We now compare our fusion model, which already outperforms all individual camera based estimates, against other state-of-the-art methods on the nuScenes dataset. 
In our comparisons, we use monocular ORB-SLAM \cite{mur2015orb}, DeepVO \cite{wang2017deepvo} and the ROS implementation of an \ac{ekf} \cite{moore2016generalized}. We use the front camera images for ORB-SLAM and DeepVO. To be able to compare ORB-SLAM in the context of monocular \ac{vo}, we run it without loop-closure following the protocols of \cite{li2018undeepvo, wang2018end}. 
Since classical monocular \ac{vo} algorithms do not recover the absolute scale for their estimated trajectories, we scale \& align ORB-SLAM trajectories with the ground truth using a least-squares similarity transform \cite{umeyama1991least}. 
For DeepVO results, we use its PyTorch implementation\footnote{https://github.com/ChiWeiHsiao/DeepVO-pytorch/} and for a fair comparison we train it with the same training split as used to train our fusion module.
In order to evaluate \ac{ekf} fusion performance, it is necessary to provide the algorithm with accurate pose and uncertainty estimates. In this test, we wish to compare our deep fusion to a traditional \ac{ekf} fusion approach. As such, we feed the \ac{mdn} outputs from each individual camera, including their estimated covariances, to the \ac{ekf}.

In Table~\ref{table:nuscenes}, we share the \ac{rpe} for three different categories based on weather/lighting conditions e.g. daylight, rain and night time. 
As a classical \ac{vo} algorithm, ORB-SLAM works by extracting and tracking salient features in consecutive frames. 
Thus, it needs sufficient illumination and texture in the environment to be able to work effectively. This explains why, in our experiments, ORB-SLAM can often not initialise successfully for night time scenarios.
Thus, we could not report its qualitative (See Figure~\ref{fig:nuscenesresult:c}) and quantitative (See Table~\ref{table:nuscenes}) analysis for night time driving conditions. 
In contrast, deep learning based approaches do produce results for night time scenarios, which is one of the most challenging conditions to estimate \ac{vo} due to the reduced visibility.

The Kalman filter is sensitive to initial parameter selection e.g. the initial estimate of covariance, process noise etc. 

We choose the set of parameters that yielded the best performance on average across the $100$ test sequences. 
Under this setup, the \ac{ekf} produces better results than ORB-SLAM. 

In the daylight and rain scenarios, the \ac{ekf} also performs better than DeepVO. However, for the night sequences, it has a larger error. This can be explained by the fact that an optimal Kalman filter requires careful parameter tuning, which is difficult to achieve for both daytime and nighttime. However, our fusion based approach outperforms best across all scenarios.

We share qualitative examples from all three categories (daylight, rain and night) in Figure~\ref{fig:nuscenesresult}. This figure shows the estimated trajectories against ground truth, and a sample image from the corresponding scene. 

As can be seen, all approaches estimate the trajectories accurately in daylight condition. However, ORB-SLAM's performance decreases significantly in rain and night time scenarios. Deep approaches maintain accurate trajectories even in these challenging rain and night-time conditions. Most significantly, our multi camera fusion approach (MC-FUSION) outperforms other methods estimating more accurate trajectories.

\section{CONCLUSION}
\label{sec:conclusion}
This paper presents a novel deep learning based multi-camera fusion framework to estimate \ac{vo} and evaluate it on the nuScenes dataset. Exploiting the advantages of deep learning, we show that the proposed approach can estimate trajectories even in night-time scenarios which is challenging for classical \ac{vo} approaches. Furthermore, we demonstrate that our proposed fusion model outperforms single camera based estimations by exploiting the complementary information from multiple camera views. Finally, we validate the performance of our approach by comparing it with the state-of-the-art methods namely, DeepVO and ORB-SLAM. We also show that the efficacy of our single camera based pose and uncertainty estimations by feeding the \ac{ekf} with them.

{\small
\bibliographystyle{./IEEETran}
\bibliography{egbib}
}

\end{document}